\documentclass[11pt,a4paper]{article}
\usepackage{times,latexsym}
\usepackage{url}
\usepackage[T1]{fontenc}

\usepackage[acceptedWithA]{tacl2018v2}

\usepackage[]{tacl2018v2}

\usepackage{booktabs}
\usepackage{CJKutf8}
\usepackage{graphicx}
\makeatletter 
  \newcommand\figcaption{\def\@captype{figure}\caption} 
  \newcommand\tabcaption{\def\@captype{table}\caption} 
\makeatother
\usepackage{xcolor}
\usepackage{multirow}

\usepackage{xspace,mfirstuc,tabulary}

\newif\iftaclinstructions
\taclinstructionsfalse % AUTHORS: do NOT set this to true
\iftaclinstructions
\renewcommand{\confidential}{}
\renewcommand{\anonsubtext}{(No author info supplied here, for consistency with
TACL-submission anonymization requirements)}
\newcommand{\instr}
\fi

\iftaclpubformat % this "if" is set by the choice of options

\else

\fi

\newcommand{\datasetName}{CrossWOZ\xspace}

\newif\ifrevision
\revisionfalse

\title{\datasetName: A Large-Scale Chinese Cross-Domain Task-Oriented Dialogue Dataset}

\author{
 Qi Zhu$^1$, Kaili Huang$^2$, Zheng Zhang$^1$,  Xiaoyan Zhu$^1$, Minlie Huang$^1$\footnotemark[1]\\
 $^1$Dept. of Computer Science and Technology,
 $^1$Institute for Artificial Intelligence, \\
 $^1$Beijing National Research Center for Information Science and Technology, \\
 $^2$Dept. of Industrial Engineering, \\Tsinghua University, Beijing, China\\
  {\sf \{zhu-q18,hkl16,z-zhang15\}@mails.tsinghua.edu.cn} \\
  {\sf \{zxy-dcs,aihuang\}@tsinghua.edu.cn}\\
}

\date{}

\begin{document}
\maketitle

\renewcommand{\thefootnote}{\fnsymbol{footnote}}
\footnotetext[1]{Corresponding author.}

\begin{abstract}
To advance multi-domain (cross-domain) dialogue modeling as well as alleviate the shortage of Chinese task-oriented datasets, we propose \textbf{\datasetName}, the first large-scale Chinese Cross-Domain Wizard-of-Oz task-oriented dataset. It contains 6K dialogue sessions and 102K utterances for 5 domains, including hotel, restaurant, attraction, metro, and taxi. Moreover, the corpus contains rich annotation of dialogue states and dialogue acts at both user and system sides. About 60\% of the dialogues have cross-domain user goals that favor inter-domain dependency and encourage natural transition across domains in conversation. We also provide a user simulator and several benchmark models for pipelined task-oriented dialogue systems, which will facilitate researchers to compare and evaluate their models on this corpus.
The large size and rich annotation of \textbf{\datasetName} make it suitable to investigate a variety of tasks in cross-domain dialogue modeling, such as dialogue state tracking, policy learning, user simulation, etc.
\end{abstract}

\section{Introduction}
\label{section1}
Recently, there have been a variety of task-oriented dialogue models thanks to the prosperity of neural architectures \cite{yao2013rnnNLU,wen2015sclstm,mrkvsic2017neuralDST,HRL,lei2018sequicity,gur2018user}. 
However, the research is still largely limited by the availability of large-scale high-quality dialogue data. Many corpora have advanced the research of task-oriented dialogue systems, most of which are single domain conversations, including ATIS \cite{ATIS1990}, DSTC 2 \cite{DSTC2}, Frames \cite{frames2017}, KVRET \cite{KVRET2017}, WOZ 2.0 \cite{Camrest2017} and M2M \cite{M2M2018}.

Despite the significant contributions to the community, these datasets are still limited in size, language variation, or task complexity.
Furthermore, there is a gap between existing dialogue corpora and real-life human dialogue data.
In real-life conversations, it is natural for humans to transition between different domains or scenarios while still maintaining coherent contexts. Thus, real-life dialogues are much more complicated than those dialogues that are only simulated within a single domain. To address this issue, some multi-domain corpora have been proposed \cite{Multiwoz2018,SchemaGuided}. 
The most notable corpus is MultiWOZ \cite{Multiwoz2018}, a large-scale multi-domain dataset which consists of crowdsourced human-to-human dialogues. It contains 10K dialogue sessions and 143K utterances for 7 domains, with annotation of system-side dialogue states and dialogue acts. However, the state annotations are noisy \cite{eric2019multiwoz21}, and user-side dialogue acts are missing. The dependency across domains is simply embodied in imposing the same pre-specified constraints on different domains, such as requiring both a hotel and an attraction to locate in the center of the town.

\begin{figure}[htbp]
    \includegraphics{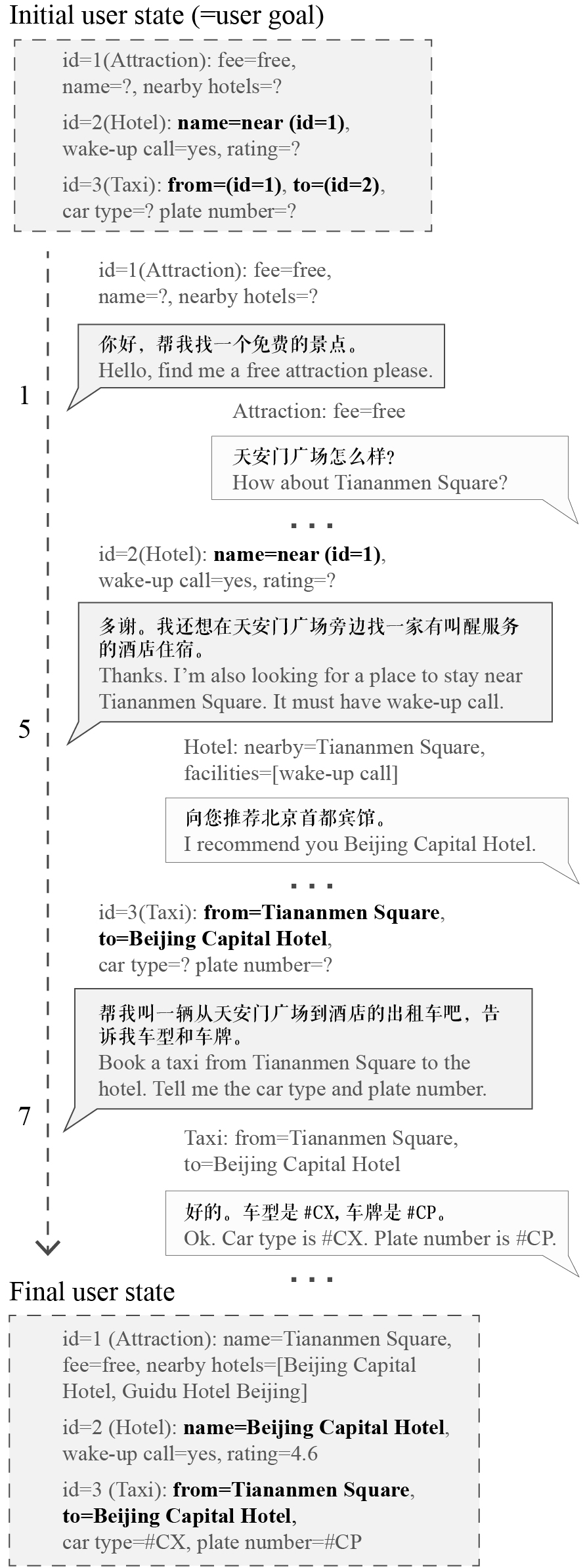}
    \caption{A dialogue example.
    The user state is initialized by the user goal: finding an attraction and one of its nearby hotels, then book a taxi to commute between these two places. 
    In addition to expressing pre-specified informable slots and filling in requestable slots, users need to consider and modify cross-domain informable slots (\textbf{bold}) that vary through conversation.
    We only show a few turns (turn number on the left), each with either user or system state of the current domain which are shown above each utterance.
    }
    \label{fig:example}
\end{figure}

In comparison to the abundance of English dialogue data, surprisingly, there is still no widely recognized Chinese task-oriented dialogue corpus. 
In this paper, we propose \textbf{\datasetName}, a large-scale Chinese  multi-domain (cross-domain) task-oriented dialogue dataset. 
An dialogue example is shown in Figure \ref{fig:example}. We compare \textbf{\datasetName} to other corpora in Table \ref{tab:cmp} and \ref{tab:cmp_case}. Our dataset has the following features comparing to other corpora (particularly MultiWOZ \cite{Multiwoz2018}):
\begin{enumerate}
    \item The dependency between domains is more challenging because the choice in one domain will affect the choices in related domains in \datasetName.
    As shown in Figure \ref{fig:example} and Table \ref{tab:cmp_case}, the hotel must be near the attraction chosen by the user in previous turns, which requires more accurate context understanding.

    \item It is the first Chinese corpus that contains large-scale multi-domain task-oriented dialogues, consisting of 6K sessions and 102K utterances for 5 domains (attraction, restaurant, hotel, metro, and taxi). 
    
    \item Annotation of dialogue states and dialogue acts is provided for both the system side and user side. The annotation of user states enables us to track the conversation from the user's perspective and can empower the development of more elaborate user simulators.
\end{enumerate}

In this paper, we present the process of dialogue collection and provide detailed data analysis of the corpus. Statistics show that our cross-domain dialogues are complicated. 
To facilitate model comparison, benchmark models are provided for different modules in pipelined task-oriented dialogue systems, including natural language understanding, dialogue state tracking, dialogue policy learning, and natural language generation. We also provide a user simulator, which will facilitate the development and evaluation of dialogue models on this corpus. 
The corpus and the benchmark models are publicly available at \url{https://github.com/thu-coai/CrossWOZ}.

\begin{table*}[ht]
\centering
\setlength{\tabcolsep}{1.5mm}{
\begin{tabular}{l|ccccc|ccc}
\toprule
\multicolumn{1}{l|}{Type} & \multicolumn{5}{c|}{Single-domain goal}           & \multicolumn{3}{c}{Multi-domain goal}                 \\ \hline
Dataset      & DSTC2  & WOZ 2.0 & Frames & KVRET  & M2M    & MultiWOZ & Schema & \textbf{\datasetName} \\
\hline
Language     & EN     & EN      & EN     & EN     & EN     & EN       & EN      & \textbf{CN}      \\
Speakers      & H2M    & H2H     & H2H    & H2H    & M2M    & H2H      & M2M     & \textbf{H2H}     \\
\# Domains   & 1      & 1       & 1      & 3      & 2      & 7        & 16      & \textbf{5}       \\
\# Dialogues & 1,612  & 600     & 1,369  & 2,425  & 1,500  & 8,438    & 16,142  & \textbf{5,012}   \\
\# Turns     & 23,354 & 4,472   & 19,986 & 12,732 & 14,796 & 115,424  & 329,964 &  \textbf{84,692}       \\
Avg. domains     & 1 & 1   & 1 & 1 & 1 &  1.80 & 1.84 &  \textbf{3.24}       \\
Avg. turns     & 14.5 & 7.5   & 14.6 & 5.3 & 9.9 & 13.7  & 20.4 &  \textbf{16.9}       \\
\# Slots     & 8      & 4       & 61     & 13     & 14     & 25       & 214     & \textbf{72}      \\
\# Values    & 212    & 99      & 3,871  & 1363   & 138    & 4,510    & 14,139  & \textbf{7,871}        \\ 
\bottomrule
\end{tabular}}
\caption{Comparison of \textbf{\datasetName}  to other task-oriented corpora (training set). H2H, H2M, M2M represent human-to-human, human-to-machine, machine-to-machine respectively. The average numbers of domains and turns are for each dialogue.
} 
\label{tab:cmp}
\end{table*}

\begin{table*}[ht]
\begin{CJK*}{UTF8}{gbsn}
\footnotesize
\setlength{\tabcolsep}{0.5mm}{
\begin{tabular}{|p{5.9cm}|p{9.9cm}|}
\hline
\multicolumn{1}{|c|}{\textbf{MultiWOZ}} & \multicolumn{1}{c|}{\textbf{\datasetName}} \\ 
\hline
usr: I'm looking for a college type attraction. & usr: 你好，可以帮我推荐一个评分是4.5分以上的景点吗？\\
\qquad ... & \qquad Hello, could you recommend an attraction with a rating of 4.5 or higher? \\
usr: I would like to visit \underline{in town centre please}. & sys: 天安门城楼，簋街小吃和\underline{北京欢乐谷}都是很不错的地方呢。\\
\qquad ... & \qquad Tiananmen, Gui Street, and \underline{Beijing Happy Valley} are very nice places. \\
usr: Can you find an Indian restaurant for me & usr: 我喜欢\underline{北京欢乐谷}，你知道这个景点周边的酒店都是什么吗？\\
that is \textbf{also in the town centre}? & \qquad I like \underline{Beijing Happy Valley}. What hotels are around this attraction?\\ \cline{1-1}
\multicolumn{1}{|c|}{\textbf{Schema}} & sys: 那可多了，有A酒店, B酒店, C酒店。\\ \cline{1-1}
usr: I want a hotel in \underline{San Diego} and I want to & \qquad There are many, such as hotel A, hotel B, and hotel C. \\
check out on Thursday next week. & usr: 太好了，我正打算在\textbf{景点附近}找个酒店住宿呢，知道哪家评分 \\
\qquad ... & 是4分以上，提供叫醒服务的不？\\
usr: I need a one way flight to \textbf{go there}. & \qquad Great! I am planning to find a hotel to stay \textbf{near the attraction}. Which one has a rating of 4 or higher and offers wake-up call service? \\ 
\hline
\end{tabular}}
\caption{
Cross-domain dialog examples in MultiWOZ, Schema, and \datasetName. The value of cross-domain constraints(\textbf{bold}) are \underline{underlined}. Some turns are omitted to save space. Names of hotels are replaced by A,B,C for simplicity. Cross-domain constraints are pre-specified in MultiWOZ and Schema, while determined dynamically in \datasetName. In \datasetName, the choice in one domain will greatly affect related domains.
}
\label{tab:cmp_case}
\end{CJK*}
\end{table*}

\section{Related Work}
\label{section2}
According to whether the dialogue agent is human or machine, we can group the collection methods of existing task-oriented dialogue datasets into three categories. The first one is \textbf{human-to-human} dialogues. One of the earliest and well-known ATIS dataset \cite{ATIS1990}  used this setting, followed by \citet{frames2017}, \citet{KVRET2017}, \citet{Camrest2017}, \citet{dealornot2017}, \citet{MedicalCH2018} and \citet{Multiwoz2018}. Though this setting requires many human efforts, it can collect natural and diverse dialogues. The second one is \textbf{human-to-machine} dialogues, which need a ready dialogue system to converse with humans. The famous Dialogue State Tracking Challenges provided a set of human-to-machine dialogue data \cite{DSTC2013,DSTC2}. The performance of the dialogue system will largely influence the quality of dialogue data. The third one is \textbf{machine-to-machine} dialogues. It needs to build both user and system simulators to generate dialogue outlines, then use templates \cite{HRL} to generate dialogues or further employ people to paraphrase the dialogues to make them more natural \cite{M2M2018,SchemaGuided}. It needs much less human effort. However, the complexity and diversity of dialogue policy are limited by the simulators. To explore dialogue policy in multi-domain scenarios, and to collect natural and diverse dialogues, we resort to the human-to-human setting. 

Most of the existing datasets only involve single domain in one dialogue, except MultiWOZ \cite{Multiwoz2018} and Schema \cite{SchemaGuided}. MultiWOZ dataset has attracted much attention recently, due to its large size and multi-domain characteristics. It is at least one order of magnitude larger than previous datasets, amounting to 8,438 dialogues and 115K turns in the training set. It greatly promotes the research on multi-domain dialogue modeling, such as policy learning \cite{takanobu2019guided}, state tracking \cite{TRADE}, and context-to-text generation \cite{MDRG}. Recently the Schema dataset is collected in a machine-to-machine fashion, resulting in 16,142 dialogues and 330K turns for 16 domains in the training set. However, the multi-domain dependency in these two datasets is only embodied in imposing the same pre-specified constraints on different domains, such as requiring a restaurant and an attraction to locate in the same area, or the city of a hotel and the destination of a flight to be the same (Table \ref{tab:cmp_case}).

Table \ref{tab:cmp} presents a comparison between our dataset with other task-oriented datasets. In comparison to MultiWOZ, our dataset has a comparable scale: 5,012 dialogues and 84K turns in the training set. The average number of domains and turns per dialogue are larger than those of MultiWOZ, which indicates that our task is more complex. The cross-domain dependency in our dataset is natural and challenging. 
For example, as shown in Table \ref{tab:cmp_case}, the system needs to recommend a hotel near the attraction chosen by the user in previous turns. Thus, both system recommendation and user selection will dynamically impact the dialogue. 
We also allow the same domain to appear multiple times in a user goal since a tourist may want to go to more than one attraction. 

To better track the conversation flow and model user dialogue policy, we provide annotation of \textbf{user states} in addition to system states and dialogue acts.
While the system state tracks the dialogue history, the user state is maintained by the user and indicates whether the sub-goals have been completed, which can be used to predict user actions. This information will facilitate the construction of the user simulator. 

To the best of our knowledge, \textbf{\datasetName} is the first large-scale Chinese dataset for task-oriented dialogue systems, which will largely alleviate the shortage of Chinese task-oriented dialogue corpora that are publicly available.

\section{Data Collection}
Our corpus is to simulate scenarios where a traveler seeks tourism information and plans her or his travel in Beijing. Domains include hotel, attraction, restaurant, metro, and taxi.
The data collection process is summarized as below: 
\begin{enumerate}
    \item \textbf{Database Construction}: we crawled travel information in Beijing from the Web, including Hotel, Attraction, and Restaurant domains (hereafter we name the three domains as HAR domains). Then, we used the metro information of entities in HAR domains to build the metro database.
    For the taxi domain, there is no need to store the information. Instead, we can call the API directly if necessary.
    
    \item \textbf{Goal Generation}: a multi-domain goal generator was designed based on the database. The relation across domains is captured in two ways. One is to constrain two targets that locate near each other. The other is to use a taxi or metro to commute between two targets in HAR domains mentioned in the context. To make workers understand the task more easily, we crafted templates to generate natural language descriptions for each structured goal. 
    
    \item \textbf{Dialogue Collection}: before the formal data collection starts, we required the workers to make a small number of dialogues and gave them feedback about the dialogue quality. Then, well-trained workers were paired to converse according to the given goals. The workers were also asked to annotate both user states and system states.
    
    \item \textbf{Dialogue Annotation}: we used some rules to automatically annotate dialogue acts according to user states, system states, and dialogue histories. 
    To evaluate the quality of the annotation of dialogue acts and states, three experts were employed to manually annotate dialogue acts and states for 50 dialogues. The results show that our annotations are of high quality.
    Finally, each dialogue contains a structured goal, a task description, user states, system states, dialogue acts, and utterances. 

\end{enumerate}

\begin{table}[b]
\centering
\setlength{\tabcolsep}{2.2mm}{
\begin{tabular}{lccc}
\toprule
  Domain & Attract.   & Rest.  & Hotel   \\
\midrule
 \# Entities & 465    & 951    & 1133    \\ 
 \# Slots & 9    & 10    & 8+37$^*$    \\ 
 Avg. nearby attract. & 4.7   & 3.3   & 0.8   \\
 Avg. nearby rest. & 6.7 & 4.1 & 2.0 \\
 Avg. nearby hotels& 2.1   & 2.4   & -  \\
\bottomrule
\end{tabular}}
\caption{Database statistics. $^*$ indicates that there are 37 binary slots for hotel services such as wake-up call. The last three rows show the average number of nearby attractions/restaurants/hotels for each entity. We did not collect nearby hotels information for the hotel domain.}
\label{tab:db}
\end{table}

\subsection{Database Construction}
We collected 465 attractions, 951 restaurants, and 1,133 hotels in Beijing from the Web. Some statistics are shown in Table \ref{tab:db}. There are three types of slots for each entity: common slots such as name and address; binary slots for hotel services such as wake-up call; nearby attractions/restaurants/hotels slots that contain nearby entities in the attraction, restaurant, and hotel domains. Since it is not usual to find another nearby hotel in the hotel domain, we did not collect such information. This nearby relation allows us to generate natural cross-domain goals, such as "find another attraction near the first one" and "find a restaurant near the attraction". Nearest metro stations of HAR entities form the metro database. In contrast, we provided the pseudo car type and plate number for the taxi domain.

\begin{table}[t]
    \centering
    \setlength{\tabcolsep}{2.25mm}{ 
    \begin{tabular}{cccc}
        \toprule
        Id & Domain & Slot & Value \\
        \midrule
        1 & Attraction & fee & \textit{free}\\
        1 & Attraction & name & \underline{\quad\quad\quad}\\
        1 & Attraction & nearby hotels & \underline{\quad\quad\quad}\\
        2 & Hotel & name & \textbf{near (id=1)} \\
        2 & Hotel & wake-up call & \textit{yes} \\
        2 & Hotel & rating & \underline{\quad\quad\quad}\\
        3 & Taxi & from & \textbf{(id=1)} \\
        3 & Taxi & to & \textbf{(id=2)} \\
        3 & Taxi & car type & \underline{\quad\quad\quad}\\
        3 & Taxi & plate number & \underline{\quad\quad\quad}\\
        \bottomrule
    \end{tabular}}
    \caption{A user goal example (translated into English). Slots with bold/italic/blank value are cross-domain informable slots, common informable slots, and requestable slots. In this example, the user wants to find an attraction and one of its nearby hotels, then book a taxi to commute between these two places.} 
    \label{tab:goal_example}
\end{table}

\subsection{Goal Generation}
\label{goal_generation}
To avoid generating overly complex goals, each goal has at most five sub-goals. To generate more natural goals, the sub-goals can be of the same domain, such as two attractions near each other. The goal is represented as a list of (sub-goal id, domain, slot, value) tuples, named as \textbf{semantic tuples}. The sub-goal id is used to distinguish sub-goals which may be in the same domain. 
There are two types of slots: informable slots which are the constraints that the user needs to inform the system, and requestable slots which are the information that the user needs to inquire from the system.
As shown in Table \ref{tab:goal_example}, besides common informable slots (italic values) whose values are determined before the conversation, we specially design cross-domain informable slots (bold values) whose values refer to other sub-goals. 
Cross-domain informable slots utilize sub-goal id to connect different sub-goals. Thus the actual constraints vary according to the different contexts instead of being pre-specified. The values of common informable slots are sampled randomly from the database. Based on the informable slots, users are required to gather the values of requestable slots (blank values in Table \ref{tab:goal_example}) through conversation.

There are four steps in goal generation. 
First, we generate independent sub-goals in HAR domains.
For each domain in HAR domains, with the same probability $\mathcal{P}$ we generate a sub-goal, while with the probability of $1-\mathcal{P}$ we do not generate any sub-goal for this domain. 
Each sub-goal has common informable slots and requestable slots. As shown in Table \ref{tab:slot_type}, all slots of HAR domains can be requestable slots, while the slots with an asterisk can be common informable slots.

Second, we generate cross-domain sub-goals in HAR domains. For each generated sub-goal (e.g., the attraction sub-goal in Table \ref{tab:goal_example}), if its requestable slots contain "nearby hotels", we generate an additional sub-goal in the hotel domain (e.g., the hotel sub-goal in Table \ref{tab:goal_example}) with the probability of $\mathcal{P}_{attraction\rightarrow hotel}$. Of course, the selected hotel must satisfy the \emph{nearby} relation to the attraction entity. 
Similarly, we do not generate any additional sub-goal in the hotel domain with the probability of $1-\mathcal{P}_{attraction\rightarrow hotel}$.
This also works for the attraction and restaurant domains. $\mathcal{P}_{hotel\rightarrow hotel}=0$ since we do not allow the user to find the nearby hotels of one hotel.

\begin{table}[t]
    \centering
    \setlength{\tabcolsep}{1.5mm}{ 
    \begin{tabular}{l}
        \toprule
        Attraction domain \\
        \textbf{name}$^*$, rating$^*$, fee$^*$, duration$^*$, address, phone, \\
        nearby attract., nearby rest., nearby hotels\\
        \midrule
        Restaurant domain \\
        \textbf{name}$^*$, rating$^*$, cost$^*$, dishes$^*$, address, phone,\\
        open, nearby attract., nearby rest., nearby hotels\\
        \midrule
        Hotel domain \\
        \textbf{name}$^*$, rating$^*$, price$^*$, type$^*$, 37 services$^*$,\\ 
        phone, address, nearby attract., nearby rest.\\
        \midrule
        Taxi domain \\
        \textbf{from}, \textbf{to}, car type, plate number\\
        \midrule
        Metro domain\\
        \textbf{from}, \textbf{to}, from station, to station\\
        \bottomrule
    \end{tabular}}
    \caption{All slots in each domain (translated into English). Slots in bold can be cross-domain informable slots. Slots with asterisk are informable slots. All slots are requestable slots except "from" and "to" slots in the taxi and metro domains. The "nearby attractions/restaurants/hotels" slots and the "dishes" slot can be multiple valued (a list). The value of each "service" is either yes or no.}
    \label{tab:slot_type}
\end{table}

Third, we generate sub-goals in the metro and taxi domains. With the probability of $\mathcal{P}_{taxi}$, we generate a sub-goal in the taxi domain (e.g., the taxi sub-goal in Table \ref{tab:goal_example}) to commute between two entities of HAR domains that are already generated. It is similar for the metro domain and we set $\mathcal{P}_{metro}=\mathcal{P}_{taxi}$.
All slots in the metro or taxi domain appear in the sub-goals and must be filled. As shown in Table \ref{tab:slot_type}, \textbf{from} and \textbf{to} slots are always cross-domain informable slots, while others are always requestable slots.

Last, we rearrange the order of the sub-goals to generate more natural and logical user goals. We require that a sub-goal should be followed by its referred sub-goal as immediately as possible.

To make the workers aware of this cross-domain feature, we additionally provide a task description for each user goal in natural language, which is generated from the structured goal by hand-crafted templates.

Compared with the goals whose constraints are all pre-specified, our goals impose much more dependency between different domains, which will significantly influence the conversation. The exact values of cross-domain informable slots are finally determined according to the dialogue context.

\subsection{Dialogue Collection}
We developed a specialized website that allows two workers to converse \emph{synchronously} and make annotations online. On the website, workers are free to choose one of the two roles: tourist (user) or system (wizard). Then, two paired workers are sent to a chatroom. The user needs to accomplish the allocated goal through conversation while the wizard searches the database to provide the necessary information and gives responses. Before the formal data collection, we trained the workers to complete a small number of dialogues by giving them feedback. Finally, 90 well-trained workers are participating in the data collection.

In contrast, MultiWOZ \cite{Multiwoz2018} hired more than a thousand workers to converse \emph{asynchronously}. Each worker received a dialogue context to review and need to respond for only one turn at a time. The collected dialogues may be incoherent because workers may not understand the context correctly and multiple workers contributed to the same dialogue session, possibly leading to more variance in the data quality. For example, some workers expressed two mutually exclusive constraints in two consecutive user turns and failed to eliminate the system's confusion in the next several turns.
Compared with MultiWOZ, our synchronous conversation setting may produce more coherent dialogues.

\subsubsection{User Side}
\label{sec:user-side}
The \textbf{user state} is the same as the user goal before a conversation starts.
At each turn, the user needs to 1) modify the user state according to the system response at the preceding turn, 2) select some semantic tuples in the user state, which indicates the dialogue acts,
and 3) compose the utterance according to the selected semantic tuples. 
In addition to filling the required values and updating cross-domain informable slots with real values in the user state, the user is encouraged to modify the constraints when there is no result under such constraints. 
The change will also be recorded in the user state. Once the goal is completed (all the values in the user state are filled), the user can terminate the dialogue.

\subsubsection{Wizard Side}
\label{sec:3.3.2}
We regard the database query as the \textbf{system state}, which records the constraints of each domain till the current turn. 
At each turn, the wizard needs to 1) fill the query according to the previous user response and search the database if necessary, 2) select the retrieved entities, and 3) respond in natural language based on the information of the selected entities. 
If none of the entities satisfy all the constraints, the wizard will try to relax some of them for a recommendation, resulting in multiple queries. The first query records original user constraints while the last one records the constraints relaxed by the system.

\subsection{Dialogue Annotation}
After collecting the conversation data, we used some rules to annotate dialogue acts automatically. Each utterance can have several dialogue acts. Each dialogue act is a tuple that consists of intent, domain, slot, and value. We pre-define 6 types of intents and use the update of the user state and system state as well as keyword matching to obtain dialogue acts. 
For the user side, dialogue acts are mainly derived from the selection of semantic tuples that contain the information of domain, slot, and value. For example, if (1, Attraction, fee, free) in Table \ref{tab:goal_example} is selected by the user, then (\textbf{Inform}, Attraction, fee, free) is labelled. If (1, Attraction, name, \underline{\quad}) is selected, then (\textbf{Request}, Attraction, name, none) is labelled. If (2, Hotel, name, near (id=1)) is selected, then (\textbf{Select}, Hotel, src\_domain, Attraction) is labelled. This intent is specially designed for the "nearby" constraint. For the system side, we mainly applied keyword matching to label dialogue acts. \textbf{Inform} intent is derived by matching the system utterance with the information of selected entities. When the wizard selects multiple retrieved entities and recommend them, \textbf{Recommend} intent is labeled. When the wizard expresses that no result satisfies user constraints, \textbf{NoOffer} is labeled. For \textbf{General} intents such as "goodbye", "thanks" at both user and system sides, keyword matching is applied.

We also obtained a binary label for each semantic tuple in the user state, which indicates whether this semantic tuple has been selected to be expressed by the user. This annotation directly illustrates the progress of the conversation.

To evaluate the quality of the annotation of dialogue acts and states (both user and system states), three experts were employed to manually annotate dialogue acts and states for the same 50 dialogues (806 utterances), 10 for each goal type (see Section \ref{sec:statistics}). 
Since dialogue act annotation is not a classification problem, we didn't use Fleiss' kappa to measure the agreement among experts.
We used dialogue act F1 and state accuracy to measure the agreement between each two experts' annotations. The average dialogue act F1 is 94.59\% and the average state accuracy is 93.55\%.
We then compared our annotations with each expert's annotations which are regarded as gold standard. The average dialogue act F1 is 95.36\% and the average state accuracy is 94.95\%, which indicates the high quality of our annotations.

\section{Statistics}
\label{sec:statistics}
After removing uncompleted dialogues, we collected 6,012 dialogues in total. The dataset is split randomly for training/validation/test, where the statistics are shown in Table \ref{tab:split}. The average number of sub-goals in our dataset is 3.24, which is much larger than that in MultiWOZ (1.80) \cite{Multiwoz2018} and Schema (1.84) \cite{SchemaGuided}. The average number of turns (16.9) is also larger than that in MultiWOZ (13.7). These statistics indicate that our dialogue data are more complex. 

According to the type of user goal, we group the dialogues in the training set into five categories:
\begin{description}
    \item[Single-domain (S)] 417 dialogues have only one sub-goal in HAR domains.
    
    \item[Independent multi-domain (M)] 1573 dialogues have multiple sub-goals (2$\sim$3) in HAR domains. However, these sub-goals do not have cross-domain informable slots.
    
    \item[Independent multi-domain + traffic (M+T)] 691 dialogues have multiple sub-goals in HAR domains and at least one sub-goal in the metro or taxi domain (3$\sim$5 sub-goals). The sub-goals in HAR domains do not have cross-domain informable slots.
    
    \item[Cross multi-domain (CM)] 1,759 dialogues have multiple sub-goals (2$\sim$5) in HAR domains with cross-domain informable slots. 
    
    \item[Cross multi-domain + traffic (CM+T)] 572 dialogues have multiple sub-goals in HAR domains with cross-domain informable slots and at least one sub-goal in the metro or taxi domain (3$\sim$5 sub-goals). 
\end{description}

\begin{table}[]
    \centering
    \setlength{\tabcolsep}{1.6mm}{
    \begin{tabular}{lccc}
    \toprule
       & Train   & Valid  & Test   \\
    \midrule
    \# Dialogues                                                           & 5,012    & 500    & 500    \\ 
    \# Turns                                                               & 84,692   & 8,458   & 8,476   \\
    \# Tokens                                                              & 1,376,033 & 137,736 & 137,427 \\
    Vocab                                                                  & 12,502   & 5,202   & 5,143  \\
    Avg. sub-goals  & 3.24    & 3.26   & 3.26   \\
    Avg. STs & 14.8 & 14.9 & 15.0\\
    Avg. turns     & 16.9   & 16.9  & 17.0  \\
    Avg. tokens         & 16.3   & 16.3  & 16.2  \\
    \bottomrule
    \end{tabular}}
    \caption{Data statistics. The average numbers of sub-goals, turns, and STs (semantic tuples) are for each dialogue. The average number of tokens is for each turn.\quad~}
    \label{tab:split}
\end{table}

The data statistics are shown in Table \ref{tab:5_type}. 
As mentioned in Section \ref{goal_generation}, we generate independent multi-domain, cross multi-domain, and traffic domain sub-goals one by one. 
Thus in terms of the task complexity, we have \textbf{S<M<CM} and \textbf{M<M+T<CM+T}, which is supported by the average number of sub-goals, semantic tuples, and turns per dialogue in Table \ref{tab:5_type}. The average number of tokens also becomes larger when the goal becomes more complex. 
About 60\% of dialogues (\textbf{M+T}, \textbf{CM}, and \textbf{CM+T}) have cross-domain informable slots.
Because of the limit of maximal sub-goals number, the ratio of dialogue number of \textbf{CM+T} to \textbf{CM} is smaller than that of \textbf{M+T} to \textbf{M}.

\textbf{CM} and \textbf{CM+T} are much more challenging than other tasks because additional cross-domain constraints in HAR domains are strict and will result in more "NoOffer" situations (i.e., the wizard finds no result that satisfies the current constraints). 
In this situation, the wizard will try to relax some constraints and issue multiple queries to find some results for a recommendation while the user will compromise and change the original goal.
The negotiation process is captured by "NoOffer rate", "Multi-query rate", and "Goal change rate" in Table \ref{tab:5_type}. 
In addition, "Multi-query rate" suggests that each sub-goal in \textbf{M} and \textbf{M+T} is as easy to finish as the goal in \textbf{S}.

\begin{figure}[h]
    \includegraphics[width=\linewidth]{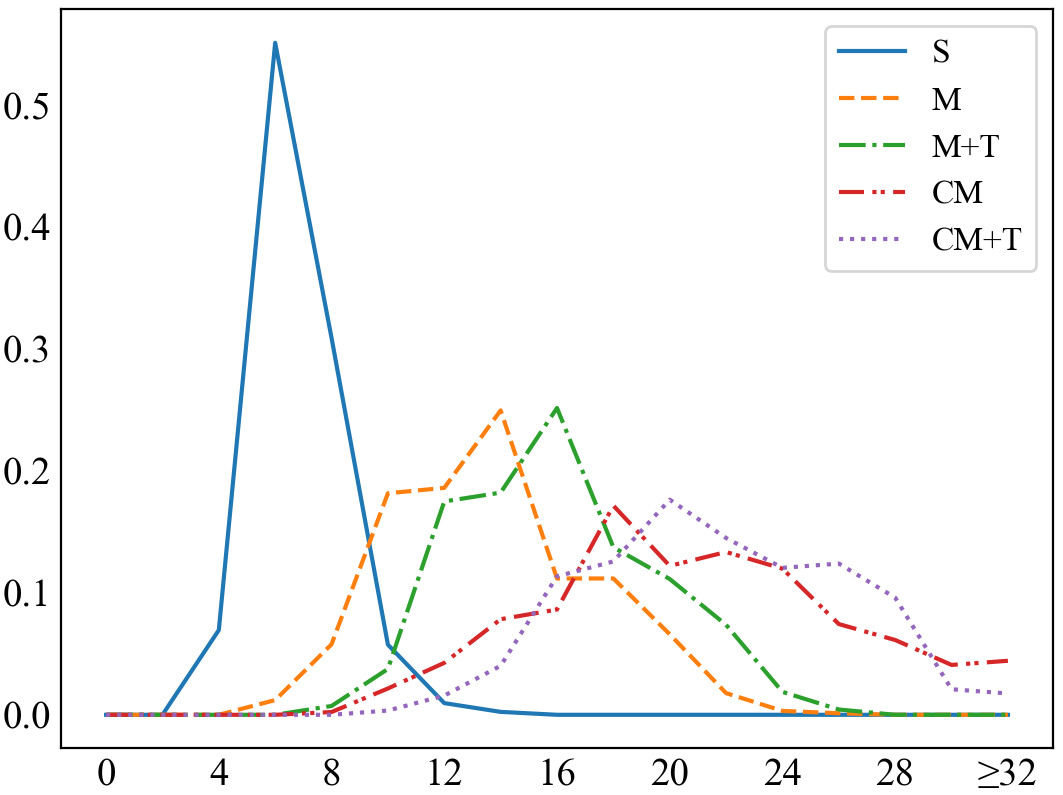}
    \caption{Distributions of dialogue length for different goal types in the training set.}
    \label{fig:turns_num_dist}
\end{figure}

The distribution of dialogue length is shown in Figure \ref{fig:turns_num_dist}, which is an indicator of the task complexity. Most single-domain dialogues terminate within 10 turns. The curves of \textbf{M} and \textbf{M+T} are almost of the same shape, which implies that the traffic task requires two additional turns on average to complete the task. The curves of \textbf{CM} and \textbf{CM+T} are less similar. This is probably because \textbf{CM} goals that have 5 sub-goals (about 22\%) can not further generate a sub-goal in traffic domains and become \textbf{CM+T} goals.

\begin{table}[t]
    \setlength{\tabcolsep}{0.7mm}{
    \begin{tabular}{lccccc}
    \toprule
        Goal type & S & M & M+T & CM & CM+T \\
    \midrule
        \# Dialogues & 417 & 1573  & 691 & 1759 & 572 \\
        NoOffer rate & 0.10 & 0.22 & 0.22 & 0.61 & 0.55\\
        Multi-query rate & 0.06 & 0.07 & 0.07 & 0.14 & 0.12 \\  
        Goal change rate & 0.10 & 0.28 & 0.31 & 0.69 & 0.63\\
        Avg. dialogue acts & 1.85 & 1.90 & 2.09 & 2.06 & 2.11\\
        Avg. sub-goals & 1.00 & 2.49 &  3.62 & 3.87 & 4.57\\
        Avg. STs & 4.5 & 11.3 & 15.8 & 18.2 & 20.7\\
        Avg. turns & 6.8 & 13.7 & 16.0 & 21.0 & 21.6\\
        Avg. tokens & 13.2 & 15.2 & 16.3 & 16.9 & 17.0\\
        
    \bottomrule
    \end{tabular}}
    \tabcaption{Statistics for dialogues of different goal types in the training set. NoOffer rate and Goal change rate are for each dialogue. Multi-query rate is for each system turn. The average number of dialogue acts is for each turn.}
    \label{tab:5_type}
\end{table}

\section{Corpus Features}
Our corpus is unique in the following aspects:
\begin{itemize}
    \item Complex user goals are designed to favor inter-domain dependency and natural transition between multiple domains. In return, the collected dialogues are more complex and natural for cross-domain dialogue tasks. 
    
    \item A well-controlled, synchronous setting is applied to collect human-to-human dialogues. This ensures the high quality of the collected dialogues.
    
    \item Explicit annotations are provided at not only the system side but also the user side. This feature allows us to model user behaviors or develop user simulators more easily.  
\end{itemize}

\section{Benchmark and Analysis}

\begin{table*}[ht]
\centering
\setlength{\tabcolsep}{1.3mm}{
\begin{tabular}{llcccccc}
\toprule
                           &                        & S     & M     & M+T   & CM    & CM+T  & Overall \\
\midrule
BERTNLU   & \multirow{2}{*}{Dialogue act F1}        & 96.69 & 96.01 & 96.15 & 94.99 & 95.38 & 95.53       \\
\ -- context & & 94.55 & 93.05 & 93.70 & 90.66 & 90.82 & 91.85 \\
\midrule
RuleDST                    & Joint state accuracy (single turn)  & 84.17      &  78.17     & 81.93      & 63.38      & 67.86      & 71.33        \\
TRADE                      & Joint state accuracy   &  71.67     & 45.29      & 37.98      &       30.77 & 25.65       & 36.08         \\
\midrule
\multirow{2}{*}{SL policy} & Dialogue act F1        & 50.28      & 44.97   & 54.01       & 41.65       & 44.02       &  44.92       \\
                          & Dialogue act F1 (delex) & 67.96      & 67.35      & 73.94      & 62.27       & 66.29       & 66.02         \\
\midrule
\multirow{2}{*}{Simulator} & Joint state accuracy (single turn)   & 63.53 & 48.79 & 50.26 & 40.66 & 41.76 & 45.00   \\
                          & Dialogue act F1 (single turn)        & 85.99      & 81.39      & 80.82      & 75.27      &  77.23     &   78.39      \\
\midrule
DA Sim  & \multirow{3}{*}{Task finish rate} & 76.5 & 49.4 & 33.7 & 17.2 & 15.7 & 34.6\\
 NL Sim (Template)  & &67.4&33.3&29.1& 10.0 & 10.0 & 23.6 \\
 NL Sim (SC-LSTM)  & &60.6&27.1&23.1& 8.8 & 9.0 & 19.7\\
\bottomrule
\end{tabular}}
\caption{Performance of Benchmark models. "Single turn" means having the gold information of the last turn. Task finish rate is evaluated on 1000 times simulations for each goal type. It's worth noting that "task finish" does not mean the task is successful, because the system may provide wrong information. Results show that cross multi-domain dialogues (\textbf{CM} and \textbf{CM+T}) is challenging for these tasks.}
\label{tab:experiments}
\end{table*}

\begin{figure}[h]
    \centering
    \includegraphics[width=0.8\linewidth]{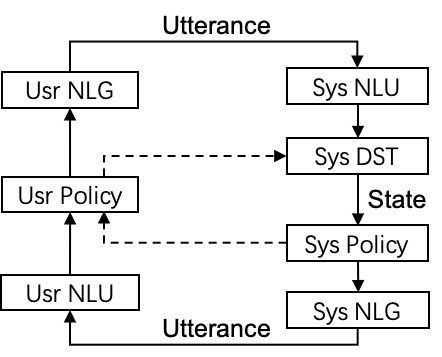}
    \caption{
    Pipelined user simulator (left) and Pipelined task-oriented dialogue system (right). Solid connections are for natural language level interaction, while dashed connections are for dialogue act level. The connections without comments represent dialogue acts.
    }
    \label{fig:framework}
\end{figure}

\textbf{\datasetName} can be used in different tasks or settings of a task-oriented dialogue system. To facilitate further research,
we provided benchmark models for different components of a pipelined task-oriented dialogue system (Figure \ref{fig:framework}), including natural language understanding (NLU), dialogue state tracking (DST), dialogue policy learning, and natural language generation (NLG). 
These models are implemented using ConvLab-2 \cite{zhu2020convlab2}, an open-source task-oriented dialog system toolkit.
We also provided a rule-based user simulator, which can be used to train dialogue policy and generate simulated dialogue data.
The benchmark models and simulator will greatly facilitate researchers to compare and evaluate their models on our corpus.

\subsection{Natural Language Understanding}
\textbf{Task}:
The natural language understanding component in a task-oriented dialogue system takes an utterance as input and outputs the corresponding semantic representation, namely, a dialogue act. The task can be divided into two sub-tasks: intent classification that decides the intent type of an utterance, and slot tagging which identifies the value of a slot.\\

\noindent\textbf{Model}:
We adapted BERTNLU from ConvLab-2. 
BERT \cite{bert2019} has shown strong performance in many NLP tasks. 
We use Chinese pre-trained BERT\footnote{BERT-wwm-ext model in \url{https://github.com/ymcui/Chinese-BERT-wwm}} \cite{chineseBERT2019}
for initialization and then fine-tune the parameters on \datasetName.
We obtain word embeddings and the sentence representation (embedding of [CLS]) from BERT. 
Since there may exist more than one intent in an utterance, we modify the traditional method accordingly. 
For dialogue acts of \emph{inform} and \emph{recommend} intents such as (intent=Inform, domain=Attraction, slot=fee, value=free) whose values appear in the sentence, we perform sequential labeling using an MLP which takes word embeddings ("free") as input and outputs tags in BIO schema ("B-Inform-Attraction-fee"). 
For each of the other dialogue acts (e.g., (intent=Request, domain=Attraction, slot=fee)) that do not have actual values, we use another MLP to perform binary classification on the sentence representation to predict whether the sentence should be labeled with this dialogue act.
To incorporate context information, we use the same BERT to get the embedding of last three utterances. We separate the utterances with [SEP] tokens and insert a [CLS] token at the beginning. 
Then each original input of the two MLP is concatenated with the context embedding (embedding of [CLS]), serving as the new input.
We also conducted an ablation test by removing context information. 
We trained models with both system-side and user-side utterances.

\begin{table}[]
    \footnotesize
    \centering
    \setlength{\tabcolsep}{0.45mm}{
    \begin{tabular}{lcccccc}
    \toprule
        & General & Inform & Request & Recom & NoOffer & Select \\
        BERTNLU & 99.45 & 94.67 & 96.57 & 98.41 & 93.87 & 82.25 \\
        \ -- context & 99.69 & 90.80 & 91.98 & 96.92 & 93.05 & 68.40 \\
        % \ -- fine-tune & 94.86 & 86.52 & 80.12 & 86.10 & 74.17 & 41.56 \\
    \bottomrule
    \end{tabular}}
    \caption{F1 score of different intent type. "Recom." represents "Recommend".}
    \label{tab:nlu_res}
\end{table}

\noindent\textbf{Result Analysis}:
The results of the dialogue act prediction (F1 score) are shown in Table \ref{tab:experiments}. We further tested the performance on different intent types, as shown in Table \ref{tab:nlu_res}. 
In general, BERTNLU performs well with context information.
The performance on cross multi-domain dialogues (\textbf{CM} and \textbf{CM+T}) drops slightly, which may be due to the decrease of "General" intent and the increase of "NoOffer" as well as "Select" intent in the dialogue data.
We also noted that the F1 score of "Select" intent is remarkably lower than those of other types, but context information can improve the performance significantly.
Since recognizing domain transition is a key factor for a cross-domain dialogue system, natural language understanding models need to utilize context information more effectively.

\subsection{Dialogue State Tracking}
\textbf{Task}:
Dialogue state tracking is responsible for recognizing user goals from the dialogue context and then encoding the goals into the pre-defined system state.
Traditional state tracking models take as input user dialogue acts parsed by natural language understanding modules, while recently there are joint models obtaining the system state directly from the context.\\

\noindent\textbf{Model}:
We implemented a rule-based model (RuleDST) and adapted TRADE (Transferable Dialogue State Generator)\footnote{\url{https://github.com/jasonwu0731/trade-dst}} \cite{TRADE} in this experiment. RuleDST takes as input the previous system state and the last user dialogue acts. Then, the system state is updated according to hand-crafted rules. 
For example, If one of user dialogue acts is (intent=Inform, domain=Attraction, slot=fee, value=free), then the value of the "fee" slot in the attraction domain will be filled with "free". 
TRADE generates the system state directly from all the previous utterances using a copy mechanism. 
As mentioned in Section \ref{sec:3.3.2}, the first query of the system often records full user constraints, while the last one records relaxed constraints for recommendation. Thus the last one involves system policy, which is out of the scope of state tracking. We used the first query for these models and left state tracking with recommendation for future work.\\

\noindent\textbf{Result Analysis}:
We evaluated the joint state accuracy (percentage of exact matching) of these two models (Table \ref{tab:experiments}).
TRADE, the state-of-the-art model on MultiWOZ, performs poorly on our dataset, indicating that more powerful state trackers are necessary.
At the test stage, RuleDST can access the previous gold system state and user dialogue acts, which leads to higher joint state accuracy than TRADE. 
Both models perform worse on cross multi-domain dialogues (\textbf{CM} and \textbf{CM+T}).
To evaluate the ability of modeling cross-domain transition, we further calculated joint state accuracy for those turns that receive "Select" intent from users (e.g., "Find a hotel near the attraction").
The performances are 11.6\% and 12.0\% for RuleDST and TRADE respectively, showing that they are not able to track domain transition well.

\subsection{Dialogue Policy Learning}
\textbf{Task}:
Dialogue policy receives state $s$ and outputs system action $a$ at each turn. Compared with the state given by a dialogue state tracker, $s$ may have more information, such as the last user dialogue acts and the entities provided by the backend database.\\

\noindent\textbf{Model}:
We adapted a vanilla policy trained in a supervised fashion from ConvLab-2 (SL policy). 
The state $s$ consists of the last system dialogue acts, last user dialogue acts, system state of the current turn, the number of entities that satisfy the constraints in the current domain, and a terminal signal indicating whether the user goal is completed. The action $a$ is delexicalized dialogue acts of current turn which ignores the exact values of the slots, where the values will be filled back after prediction.\\

\noindent\textbf{Result Analysis}:
As illustrated in Table \ref{tab:experiments}, there is a large gap between F1 score of exact dialogue act and F1 score of delexicalized dialogue act, which means we need a powerful system state tracker to find correct entities.
The result also shows that cross multi-domain dialogues (\textbf{CM} and \textbf{CM+T}) are harder for system dialogue act prediction. 
Additionally, when there is "Select" intent in preceding user dialogue acts, the F1 score of exact dialogue act and delexicalized dialogue act are 41.53\% and 54.39\% respectively. This shows that the policy performs poorly 
for cross-domain transition.

\subsection{Natural Language Generation}
\textbf{Task}:
Natural language generation transforms a structured dialogue act into a natural language sentence. It usually takes delexicalized dialogue acts as input and generates a template-style sentence that contains placeholders for slots. Then, the placeholders will be replaced by the exact values, which is called lexicalization.\\

\noindent\textbf{Model}:
We provided a template-based model (named TemplateNLG) and SC-LSTM (Semantically Conditioned LSTM) \cite{wen2015sclstm}  for natural language generation. For TemplateNLG, we extracted templates from the training set and manually added some templates for infrequent dialogue acts. For SC-LSTM we adapted the implementation\footnote{\url{https://github.com/andy194673/nlg-sclstm-multiwoz}} on MultiWOZ and trained two SC-LSTM with system-side and user-side utterances respectively.\\ 

\noindent\textbf{Result Analysis}:
We calculated corpus-level BLEU as used by \citet{wen2015sclstm}. We took all utterances with the same delexcalized dialogue acts as references (100 references on average), which results in high BLEU score.
For user-side utterances, the BLEU score for TemplateNLG is 0.5780, while the BLEU score for SC-LSTM is 0.7858. For system-side, the two scores are 0.6828 and 0.8595.
As exemplified in Table \ref{tab:nlg_case}, the gap between the two models can be attributed to that SC-LSTM generates common pattern while TemplateNLG retrieves 
original sentence which has more specific information.
We do not provide BLEU scores for different goal types (namely, \textbf{S}, \textbf{M}, \textbf{CM}, etc.) because BLEU scores on different corpus are not comparable.

\begin{table}[]
\begin{CJK*}{UTF8}{gbsn}
    \small
    \centering
    \setlength{\tabcolsep}{0.6mm}{
    \begin{tabular}{p{7.5cm}}
    \toprule
        Input: \\
        (Inform, Restaurant, name, \$name)\\ 
        (Inform, Restaurant, cost, \$cost)\\
    \midrule
        SC-LSTM: \\
        为您推荐\$name，人均消费\$cost。\\
        I Recommend you \$name. It costs \$cost. \\
    \midrule
        TemplateNLG: \\
        1) \$name是个不错的选择，但是它的人均消费是\$cost。\\
        $\quad $ \$name is a nice choice. But it costs \$cost. \\
        2) 您想吃的菜不需要花那么多钱呢。为您推荐\$name，人均消费\$cost哟。\\
        $\quad $ The dish you want doesn't cost so much. I recommend you \$name. It costs \$cost.\\
    \bottomrule
    \end{tabular}}
    \caption{
    Comparison of SC-LSTM and TemplateNLG. The input is delexicalized dialogue acts, where the actual values are replaced with \$name and \$cost. Two retrieved results are shown for TemplateNLG.
    }
    \label{tab:nlg_case}
\end{CJK*}
\end{table}

\subsection{User Simulator}

\textbf{Task}:
A user simulator imitates the behavior of users, which is useful for dialogue policy learning and automatic evaluation. 
A user simulator at dialogue act level (e.g., the "Usr Policy" in Figure \ref{fig:framework}) receives the system dialogue acts and outputs user dialogue acts, while a user simulator at natural language level (e.g., the left part in Figure \ref{fig:framework}) directly takes system's utterance as input and outputs user's utterance.\\

\noindent\textbf{Model}:
We built a rule-based user simulator that works at dialogue act level. Different from agenda-based \cite{agenda2007schatzmann} user simulator that maintains a stack-like agenda, our simulator maintains the user state straightforwardly (Section \ref{sec:user-side}). 
The simulator will generate a user goal as described in Section \ref{goal_generation}. 
At each user turn, the simulator receives system dialogue acts, modifies its state, and outputs user dialogue acts according to some hand-crafted rules.
For example, if the system inform the simulator that the attraction is free, then the simulator will fill the "fee" slot in the user state with "free", and ask for the next empty slot such as "address".
The simulator terminates when all requestable slots are filled, and all cross-domain informable slots are filled by real values. \\

\noindent\textbf{Result Analysis}:
During the evaluation, we initialized the user state of the simulator using the previous gold user state. The input to the simulator is the gold system dialogue acts. We used joint state accuracy (percentage of exact matching) to evaluate user state prediction and F1 score to evaluate the prediction of user dialogue acts. 
The results are presented in Table \ref{tab:experiments}. 
We can observe that the performance on complex dialogues (\textbf{CM} and \textbf{CM+T}) is remarkably lower than that on simple ones (\textbf{S}, \textbf{M}, and \textbf{M+T}). This simple rule-based simulator is provided to facilitate dialogue policy learning and automatic evaluation, and our corpus supports the development of more elaborated simulators as we provide the annotation of user-side dialogue states and dialogue acts.

\subsection{Evaluation with User Simulation}

In addition to corpus-based evaluation for each module, we also evaluated the performance of a whole dialogue system using the user simulator as described above. Three configurations were explored:
\begin{description}
    \item[DA Sim] Simulation at dialogue act level. As shown by the dashed connections in Figure \ref{fig:framework}, we used the aforementioned simulator at the user side and assembled the dialogue system with RuleDST and SL policy.
    
    \item[NL Sim (Template)] Simulation at natural language level using TemplateNLG. As shown by the solid connections in Figure \ref{fig:framework}, the simulator and the dialogue system were equipped with BERTNLU and TemplateNLG additionally.
    
    \item[NL Sim (SC-LSTM)] Simulation at natural language level using SC-LSTM. TemplateNLG was replaced with SC-LSTM in the second configuration.
\end{description}

When all the slots in a user goal are filled by real values, the simulator terminates. This is regarded as "task finish". It's worth noting that "task finish" does not mean the task is success, because the system may provide wrong information. We calculated "task finish rate" on 1000 times simulations for each goal type (See Table \ref{tab:experiments}). 
Findings are summarized below:

\begin{enumerate}
    \item Cross multi-domain tasks (\textbf{CM} and \textbf{CM+T}) are much harder to finish. 
    Comparing \textbf{M} and \textbf{M+T}, although each module performs well in traffic domains, additional sub-goals in these domains are still difficult to accomplish.

    \item The system-level performance is largely limited by RuleDST and SL policy.
    Although the corpus-based performance of NLU and NLG modules is high, the two modules
    still harm the performance. Thus more powerful models are needed for all components of a pipelined dialogue system.
    
    \item TemplateNLG has a much lower BLEU score but performs better than SC-LSTM in natural language level simulation. This may be attributed to that BERTNLU prefers templates retrieved from the training set.

\end{enumerate}

\section{Conclusion}
In this paper, we present the first large-scale Chinese Cross-Domain task-oriented dialogue dataset, \textbf{\datasetName}. 
It contains 6K dialogues and 102K utterances for 5 domains, with the annotation of dialogue states and dialogue acts at both user and system sides.
About 60\% of the dialogues have cross-domain user goals, which encourage natural transition between related domains.
Thanks to the rich annotation of dialogue states and dialogue acts at both user side and system side, this corpus provides a new testbed for a wide range of tasks to investigate cross-domain dialogue modeling, such as dialogue state tracking, policy learning, etc.
Our experiments show that the cross-domain constraints are challenging for all these tasks. The transition between related domains is especially challenging to model.
Besides corpus-based component-wise evaluation, we also performed system-level evaluation with a user simulator, which requires more powerful models for all components of a pipelined cross-domain dialogue system.

\section*{Acknowledgments}
This work was supported by the National Science Foundation of China (Grant No. 61936010/61876096) and the National Key R\&D Program of China (Grant No. 2018YFC0830200). 
We would like to thank THUNUS NExT JointLab for the support. 
We would also like to thank Ryuichi Takanobu and Fei Mi for their constructive comments. We are grateful to our action editor, Bonnie Webber, and the anonymous reviewers for their valuable suggestions and feedback.

\bibliography{tacl2018}
\bibliographystyle{acl_natbib}

\end{document}